\documentclass[10pt,twocolumn,letterpaper]{article}

\usepackage[pagenumbers]{cvpr}
\usepackage[utf8]{inputenc}
\usepackage{newunicodechar}
\newunicodechar{：}{:}
\usepackage{amsmath,amssymb}
\usepackage{booktabs}
\usepackage{multirow}
\usepackage{graphicx}
\usepackage{array}
\usepackage{booktabs}
\usepackage[normalem]{ulem}
\usepackage{tabularx}
\usepackage{caption}
\usepackage[table]{xcolor}
\newcommand{\grayrow}{\rowcolor{gray!20}}

\definecolor{cvprblue}{rgb}{0.21,0.49,0.74}
\usepackage[pagebackref,breaklinks,colorlinks,allcolors=cvprblue]{hyperref}

\def\confName{CVPR}
\def\confYear{2027}

\title{ReflectScene：MLLM-Guided Closed-Loop Diffusion for 3D Indoor Scene Generation}

\author{
    Jiawei Shen$^{1}$\quad Jie Feng$^{1}$\thanks{Corresponding Author.}\quad Junpeng Zhang$^{1}$\quad Guanbin Li$^{2,4}$\\
    Wanyi Chen$^3$\quad Weisheng Dong$^{1}$\quad  Guangming Shi$^{1}$\\
    {$^1$Xidian University, $^2$Sun Yat-sen University}, $^3$Soochow University  \\
    $^4$Shenzhen Loop Area Institute \\
    \\
}
\begin{document}
\twocolumn[{
  \maketitle
  \begin{center}
    \vspace{-0.7cm}
    \includegraphics[width=0.82\linewidth]{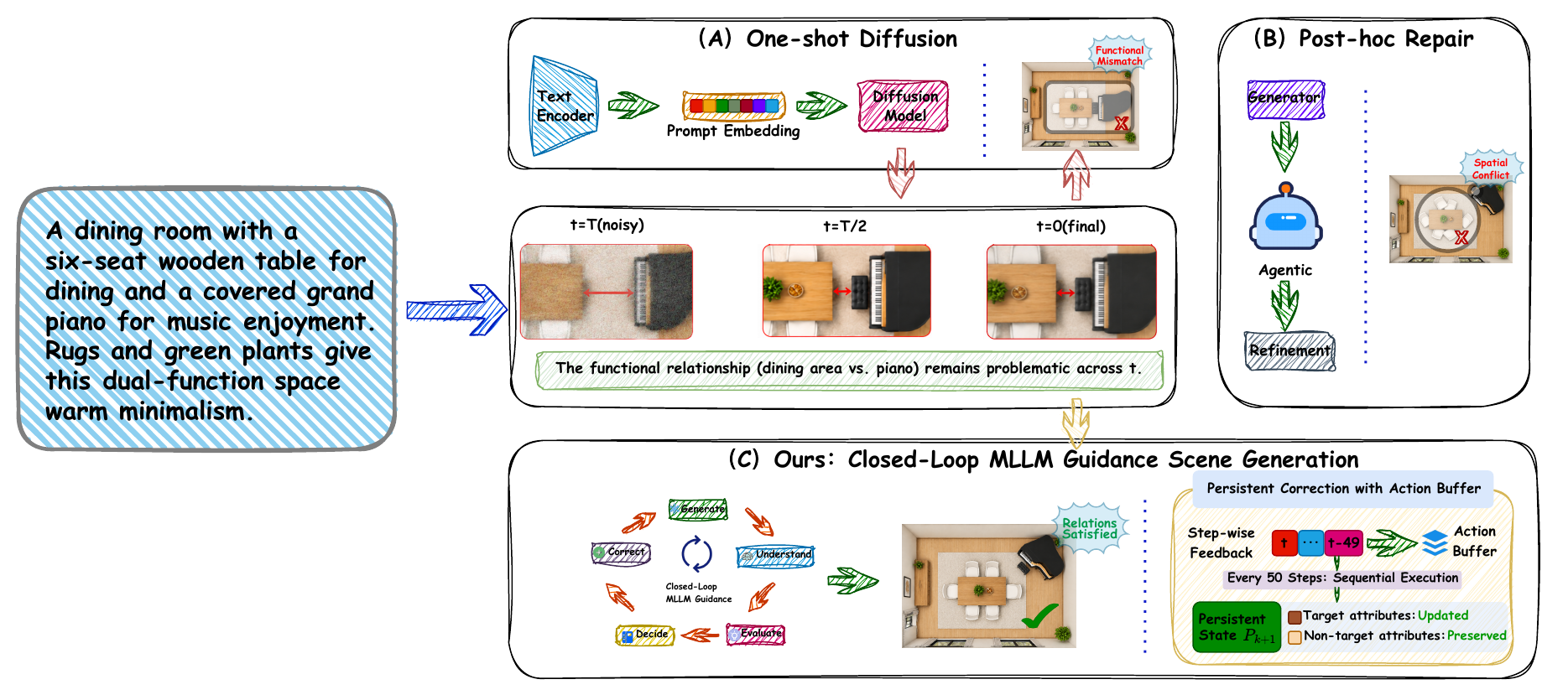}
    \vspace{-0.3cm}
     \captionof{figure}{\textbf{Motivation of ReflectScene.}
(A) One-shot diffusion relies on fixed text conditioning, lacking feedback to correct unsatisfied relations.
(B) Post-hoc repair acts after generation and may disrupt existing layouts.
(C) ReflectScene converts step-wise semantic assessments into executable object-level corrections, buffering and sequentially executing decisions every 50 steps to guide subsequent denoising while preserving non-target arrangements.}
    \label{fig:1}
    \vspace{-6pt}
  \end{center}
}]

% \begin{figure*}[!t]
% \centering
% \includegraphics[width=0.83\linewidth]{sec/image/motivation2.pdf}
% \vspace{-12pt}
% \caption{\textbf{Motivation of ReflectScene.} (A) One-shot diffusion relies on fixed prompt embeddings without explicitly understanding intermediate scene states, potentially leading to functional mismatches. (B) Post-hoc agentic refinement operates after generation and may introduce spatial conflicts. (C) Our ReflectScene establishes a closed-loop MLLM-guided generation paradigm, integrating scene generation, understanding, evaluation, decision-making, and correction to progressively improve scene coherence.}
% \label{fig:1}
% \vspace{-18pt}
% \end{figure*}

\maketitle
\begin{abstract}
Text-conditioned 3D indoor scene generation requires layouts satisfying geometric, functional, and semantic constraints. 
However, existing diffusion methods rely primarily on static textual conditions, limiting their ability to assess evolving scenes and translate semantic discrepancies into executable object-level corrections. 
MLLM-based post-hoc refinement provides semantic feedback but remains decoupled from denoising, potentially disrupting scene coherence.
We propose ReflectScene, an MLLM-guided closed-loop diffusion framework. 
First, we introduce Hierarchical Prior-Grounded Scene Understanding that aligns structured layouts with multi-view visual evidence, to interpret object-specific context and scene global semantics via a frozen multi-modal large language model (MLLM). 
Second, we propose MLLM-Guided Semantic Reward Evaluation to distill MLLM's diagnoses and assessments on possible candidate corrections into a lightweight Reward Predictor, enabling correction decision making with reduced computation burden. 
Finally, Decision-Conditioned Trajectory-Consistent Diffusion is proposed to integrate selected corrections through transient conditioning and persistent clean-estimate projection, correcting object layout in the generated scene while preserving non-target arrangements throughout denoising.
Experiments on 3D-FRONT demonstrate improved functional consistency and semantic alignment while maintaining competitive generation quality and diversity.
\end{abstract}

\section{Introduction}
\label{sec:intro}

Text-conditioned 3D indoor scene generation aims to synthesize realistic furniture layouts from natural language, with applications in virtual reality, game development, and embodied AI \cite{li2019grains,wang2021sceneformer,feng2023layoutgpt}. 
Beyond geometric plausibility, realistic scenes must satisfy functional and semantic requirements, including meaningful spatial relationships and contextual dependencies \cite{li2019grains,zhou2019scenegraphnet}.

As illustrated in \cref{fig:1}, recent learning-based approaches have achieved promising results in indoor scene synthesis. DiffuScene \cite{tang2024diffuscene} employs denoising diffusion to jointly model furniture attributes, while InstructScene \cite{lin2024instructscene} incorporates semantic graph priors for controllable generation. 
These methods primarily rely on static textual conditions or predefined structural priors, and present limited ability to reassess intermediate denoising states and correct emerging semantic discrepancies. 
Consequently, generated layouts may remain geometrically plausible while failing to satisfy the functional and contextual intent of the input description.

Meanwhile, multimodal foundation models provide rich semantic and functional knowledge for scene understanding and reasoning \cite{zhu20233d,jia2024sceneverse,chen2024ll3da}. 
Function2Scene \cite{wang2026function2scene} combines geometric measurements, LLM reasoning, and VLM assessment, while SceneReVis \cite{zhao2026scenerevis} employs vision-grounded diagnosis and executable actions for iterative scene refinement. 
Despite these advances, semantic reasoning and diffusion-based generation remain largely decoupled: static conditions cannot respond to evolving semantic discrepancies, while post-hoc corrections are not inherently propagated through subsequent denoising. 
Bridging this gap requires reliable intermediate scene understanding, context-aware evaluation of candidate corrections, and persistent integration of accepted object-level decisions into the generative trajectory, without unnecessarily disrupting satisfactory scene content. 
Moreover, repeatedly invoking large MLLMs to evaluate multiple candidates introduces substantial computational overhead.

In this paper, we present ReflectScene, an MLLM-guided closed-loop diffusion framework that integrates semantic understanding, reward evaluation, object-level decision-making, and corrective generation into a single reverse-diffusion trajectory.  
To achieve this goal, it requires faithfully representing the evolving layout,
efficiently evaluating context-dependent candidate corrections, and integrating
selected corrections into the reverse-diffusion trajectory without disrupting
satisfactory scene content.
To represent the intermediate layout, we develop Hierarchical Prior-Grounded Scene Understanding, which combines structural-semantic alignment, multi-view visual supervision, and geometry-aware metric learning to produce object-specific context and global scene representations.
To make candidate evaluation tractable, we introduce MLLM-guided Semantic Reward Evaluation and Object-Level Decision, which distills a frozen MLLM evaluator into a lightweight Reward Predictor that estimates prompt alignment, functional coherence, and semantic consistency while retaining candidate-specific scene understanding. 
To maintain the effect of selected corrections during generation, we develop Decision-conditioned Trajectory-consistent Diffusion, which feeds
accepted decisions back into the reverse-diffusion trajectory through
transient guidance and persistent correction constraints, while preserving
non-target attributes.
These components are interleaved at denoising steps, and corrections accumulated within each buffering interval are then executed sequentially, enabling progressive refinement along a single reverse-diffusion trajectory.

Our main contributions are summarized as follows:
\begin{itemize}
    \item We introduce ReflectScene that integrates semantic assessment and executable object-level corrections into a single reverse-diffusion trajectory, bridging the gap between static conditional generation and decoupled post-hoc refinement.

    \item We develop hierarchical scene understanding and MLLM-guided reward mechanism that combines structural, semantic, and metric representations with reward distillation, enabling efficient evaluation and context-aware selection of geometrically feasible corrections.

    \item We propose a Decision-conditioned Trajectory-consistent Diffusion mechanism that combines transient decision conditioning and decision-effect consistency learning to complete selected corrections while preserving non-target scene content.
\end{itemize}

\section{Related Work}
\label{sec:formatting}
%-------------------------------------------------------------------------
\subsection{3D Indoor Scene Synthesis}
3D indoor scene synthesis aims to generate plausible furniture arrangements with semantic coherence and spatial consistency. 
Early approaches primarily focus on learning object-centric scene distributions. 
For example, ATISS \cite{paschalidou2021atiss} models indoor scenes as unordered object sets using autoregressive transformers, while DiffuScene \cite{tang2024diffuscene} employs denoising diffusion to jointly generate object semantics, spatial attributes, and geometric features. 
Such generative representations\cite{zhai2023commonscenes,huang2023diffusion}  enable flexible modeling of diverse furniture configurations and scene compositions.

Recent studies\cite{feng2025casagpt,zhang2026m3dlayout,miao2026nestwork} further incorporate structural priors to improve layout plausibility and controllability. 
For instance, InstructScene \cite{lin2024instructscene} introduces semantic graph priors to capture object relationships, while MMGDreamer \cite{yang2025mmgdreamer} employs mixed-modality graphs for geometry-controllable generation. 
SemLayoutDiff \cite{sun2026semlayoutdiff} further incorporates architectural constraints into semantic layout diffusion. 
These advances \cite{yang2024holodeck,ran2025directlayout,deng2025global,feng2026lavigen} enhance the modeling of object interactions and spatial dependencies. Nevertheless, these approaches largely rely on learned distributions or predefined priors, with limited explicit feedback on evolving scene configurations.

%-------------------------------------------------------------------------
\subsection{Feedback-Guided Scene Refinement}
Improving spatial and functional plausibility has motivated various guidance and refinement strategies.
One direction incorporates explicit physical constraints into generation. 
For example, PhyScene \cite{yang2024physcene} introduces physics- and interactivity-based guidance into diffusion sampling to reduce collisions and improve spatial feasibility. 
Such guidance\cite{sun2025layoutvlm,jiang2026hoglayout} provides explicit objectives for encouraging physically consistent scene configurations.

Another direction explores reasoning-driven scene refinement through iterative feedback. 
For instance, SDesc3D\cite{feng2026sdesc3d} combines multi-view structural priors and functionality-aware layout grounding with iterative reflection-rectification to resolve spatial inconsistencies in scenes generated from short descriptions. 
Similarly, SceneReVis \cite{zhao2026scenerevis} employs vision-grounded diagnosis and corrective actions for iterative scene refinement, emphasizing the role of contextual reasoning in layout optimization\cite{kuang2026vulcan}. 
However, converting iterative feedback into persistent object-level corrections without disrupting unaffected scene content remains challenging.

%-------------------------------------------------------------------------
\subsection{Multimodal Priors for 3D Understanding and Generation}
Pretrained multimodal models offer rich semantic knowledge, motivating efforts\cite{xue2023ulip,peng2023openscene,jatavallabhula2023conceptfusion,takmaz2023openmask3d} to connect 3D representations with language models. 
For example, PointLLM \cite{xu2024pointllm} aligns point-cloud features with language representations for 3D object understanding, while 3D-LLaVA \cite{deng20253d} introduces a dedicated 3D connector for scene-level multimodal reasoning. These approaches\cite{chen2020scanrefer,chen2021scan2cap,azuma2022scanqa} facilitate the integration of geometric information with pretrained semantic knowledge.

Beyond understanding, pretrained multimodal priors have also been explored for generative modeling. 
For instance, RepFusion \cite{pan2026repfusion} incorporates MLLM priors into diffusion-based image generation through dynamic conditioning on evolving visual representations, highlighting the potential of state-dependent multimodal guidance. 
However, translating multimodal semantic assessments into precise and executable geometric corrections remains underexplored.

%-------------------------------------------------------------------------
\begin{figure*}[!t]
\centering
\includegraphics[width=0.80\linewidth]{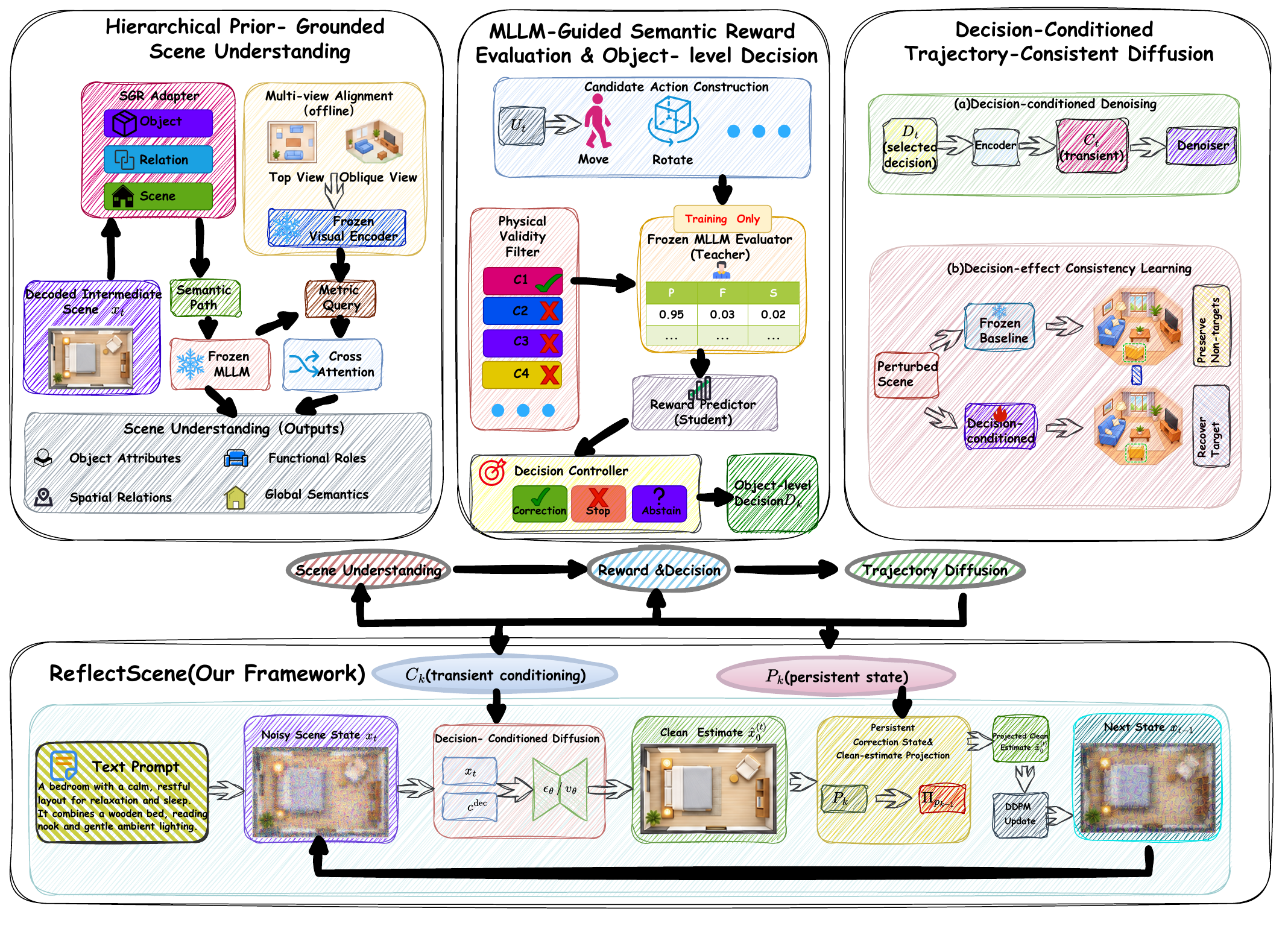}
\vspace{-12pt}
\caption{\textbf{Overview of ReflectScene.}
ReflectScene integrates hierarchical scene understanding
and MLLM-guided decisions into object-centric diffusion.
At each denoising step, intermediate clean estimates are
assessed to select object-level actions, providing transient
conditioning, while buffered non-STOP actions are executed
every 50 steps to update a persistent correction state.
This state is projected onto subsequent clean estimates
before DDPM posterior sampling, with decision-effect
consistency learning promoting target recovery and
non-target preservation.}
\label{fig:2}
\vspace{-18pt}
\end{figure*}

\section{Method}

ReflectScene formulates text-conditioned 3D scene generation as a closed-loop
process, in which intermediate layouts are interpreted, corrected and fed back
into the same reverse-diffusion trajectory. 
To achieve this goal, ReflectScene is composed of three interconnected
components: Hierarchical Prior-Grounded Scene Understanding, MLLM-guided
Semantic Reward Evaluation and Object-Level Decision, and Decision-Conditioned
Trajectory-Consistent Diffusion.

As illustrated in \cref{fig:2}, given a text prompt and an initial noise state, the diffusion model progressively generates a scene layout. 
At each denoising step checkpoint, the Hierarchical Prior-Grounded Scene Understanding module constructs structured object-specific context representations and global scene representation of the intermediate layout. 
These hierarchical representations are then passed to MLLM-guided Semantic Reward Evaluation and Object-Level Decision to evaluate candidate
corrections and select an object-level correction. 
The Decision-conditioned Trajectory-consistent Diffusion module then incorporates the selected decision into subsequent denoising while preserving accepted corrections along the reverse-diffusion trajectory.

\subsection{Hierarchical Prior-Grounded Scene Understanding}
\label{sec:understand}

At each selected denoising step, ReflectScene evaluates the intermediate layout before selecting an object-level correction, requiring a representation that jointly captures object semantics and geometry and the semantic priors needed for
functional evaluation. 
To this end, we introduce a Hierarchical Prior-Grounded Scene Understanding module, which first constructs a set of hierarchical representations of the intermediate layout and then adapts it to a frozen MLLM through complementary semantic and metric feature paths. 
The resulting representation supports candidate-specific reward evaluation and object-level decision making. 
Moreover, a multi-view alignment pretraining scheme is introduced for facilitating the adaptation of scene representations to MLLM's hidden space.

\noindent\textbf{Hierarchical Scene Representation.}
For each object $i$, we combine its category embedding $\mathbf{E}(c_i)$,
geometric attributes $\mathbf{g}_i$, and encoded diffusion timestep
$\mathbf{PE}(t)$ to obtain an initial object feature,
\begin{equation}
\mathbf{h}_i =
\mathbf{W}_c \mathbf{E}(c_i)
+ \mathbf{W}_g \mathbf{g}_i
+ \mathbf{W}_t \mathbf{PE}(t)
\in \mathbb{R}^{128},
\end{equation}
where $\mathbf{W}_c$, $\mathbf{W}_g$, and $\mathbf{W}_t$ denote three independent linear projections applied on $\mathbf{E}(c_i)$, $\mathbf{g}_i$ and $\mathbf{PE}(t)$.

Meanwhile, for each object pair $(i,j)$, we construct a pairwise relation descriptor
from their geometric attributes:
\begin{equation}
\mathbf{u}_{ij} =
\left[
\Delta\mathbf{p}_{ij},
d_{ij}^{xz},
\mathbf{b}_i,
\mathbf{b}_j,
\sin\Delta\theta_{ij},
\cos\Delta\theta_{ij}
\right]
\end{equation}
where $\Delta\mathbf{p}_{ij}$ is the 3D relative displacement,
$d_{ij}^{xz}$ is the horizontal distance, $\mathbf{b}_i,\mathbf{b}_j
\in\mathbb{R}^3$ are the object sizes, and
$\Delta\theta_{ij}=\theta_j-\theta_i$ is their relative yaw angle.

The short-range context for each object $\mathbf{z}_i^{\mathrm{short}}$ is aggregated from these pairwise descriptors by using masked attention:
\begin{equation}
\mathbf{z}_i^{\mathrm{short}} =
\operatorname{Attn}\left(
\mathbf{h}_i,\,
\left\{\operatorname{MLP}_r(\mathbf{u}_{ij})\right\}_{j=1}^{N_k},\,
\mathbf{M}_i^{\mathrm{nbr}}
\right),
\end{equation}
where $\mathbf{M}_i^{\mathrm{nbr}}$ masks all objects except the up to eight
nearest neighbors within a 2-meter radius of object $i$.

To capture long-range interactions and scene-level context, we prepend a
learnable global query $\mathbf{q}_g$ to the sequence of object features and
process the sequence with a Transformer:
\begin{equation}
\left(
\{\mathbf{z}_i^{\mathrm{long}}\}_{i=1}^{N_k},
\mathbf{z}^{\mathrm{global}}
\right)
=
\psi\left(
\mathbf{q}_g,\,
\left\{
[\mathbf{h}_i \Vert \mathbf{z}_i^{\mathrm{short}}]
\right\}_{i=1}^{N_k}
\right),
\label{eq:sgr}
\end{equation}
where $\psi(\cdot)$ is a two-layer Transformer with hidden dimension 128, and $[\cdot\Vert\cdot]$ denotes feature concatenation. 
The Transformer outputs object-specific long-range features $\mathbf{z}_i^{\mathrm{long}}$ and the global scene representation $\mathbf{z}^{\mathrm{global}}$.

\noindent\textbf{Prior-Grounded Scene Understanding.}
To incorporate the hierarchical scene representation into the frozen MLLM, we project the obtained global scene feature $\mathbf{z}^{\mathrm{global}}$ and object-level long-range features $\{\mathbf{z}_i^{\mathrm{long}}\}_{i=1}^{N_k}$, and append the resulting continuous scene tokens to the prompt embeddings.
Let $\mathbf{E}_{\mathrm{txt}}(p_s)$ denote the prompt token embeddings obtained from the MLLM's input embedding layer. 
The joint text-scene sequence is constructed as
$\mathbf{T}_{\mathrm{sem}} =
[\mathbf{E}_{\mathrm{txt}}(p_s);
\mathbf{W}_s\mathbf{z}^{\mathrm{global}};
\{\mathbf{W}_s\mathbf{z}_i^{\mathrm{long}}\}_i^{N_k}]$,
where $\mathbf{W}_s$ is a learnable projection that maps the scene
features into the 2048-dimensional hidden space of the MLLM.
After providing this combined sequence to a frozen MLLM, we extract the final-layer hidden states at the positions of the appended scene tokens as
\begin{equation}
\left(
\mathbf{h}_g^L,
\left\{
\mathbf{h}_i^L
\right\}_{i=1}^{N_k}
\right)
=
\operatorname{MLLM}_{\phi,\mathrm{hidden}}^{(L)}
\left(
\mathbf{T}_{\mathrm{sem}}
\right),
\label{eq:mllm_semantic_output}
\end{equation}
where $L$ denotes the final MLLM layer.
The resulting $\mathbf{h}_g^L$ and $\mathbf{h}_i^L$ are the MLLM-adapted global scene feature and the corresponding MLLM-adapted feature for object $i$.

To further encode object geometry in the MLLM feature space, we introduce a Metric Adapter that transforms the geometric attributes $\mathbf{g}_i$ into a geometry-aware query:
\begin{equation}
\mathbf{q}_i^m =
\mathbf{W}_m\mathbf{g}_i
+
\operatorname{MLP}_m(\mathbf{g}_i),
\label{eq:metric_query}
\end{equation}
where $\mathbf{W}_m$ and $\operatorname{MLP}_m$ are learnable branches whose outputs have the same dimensionality as the MLLM hidden states.
To incorporate the information encoded in the semantic token sequence, we
cache the keys and values produced by the first layer of the frozen MLLM:
\begin{equation}
\begin{aligned}
& (\mathbf{K}_{\mathrm{sem}},\mathbf{V}_{\mathrm{sem}})
=
\operatorname{KV}_{\phi_1}^{\mathrm{frozen}}
\left(
\mathbf{T}_{\mathrm{sem}}
\right),\\
& \mathbf{h}_i^m
=
\operatorname{CrossAttn}_{\phi_1}^{\mathrm{frozen}}
\left(
\mathbf{q}_i^m,\,
\operatorname{sg}(\mathbf{K}_{\mathrm{sem}}),\,
\operatorname{sg}(\mathbf{V}_{\mathrm{sem}})
\right),
\end{aligned}
\label{eq:mllm_geometric_output}
\end{equation}
where $\operatorname{sg}(\cdot)$ denotes the stop-gradient operation. 
The resulting metric feature $\mathbf{h}_i^m$ captures the geometry of object $i$ through interaction with the MLLM-adapted scene context.

Overall, the scene understanding module produces a hierarchical representation by combining object-level semantic and metric features with global scene semantics, which is formulated as
\begin{equation}
\label{eq:final_hierarchical_reps}
\left(
\left\{
(\mathbf{h}_i^L,\mathbf{h}_i^m)
\right\}_{i=1}^{N_k},
\mathbf{h}_g^L
\right)
=\mathcal{U}(\mathcal{S}_k,p_s).
\end{equation}
The resulting features serve as the hierarchical scene representations for subsequent reward evaluation and object-level decision making.

\noindent\textbf{Multi-view Alignment Pretraining.}
To provide visual supervision for the structured scene representation during
training, we render each scene from a set $\mathcal{V}$ of top-down and
oblique viewpoints. Each rendered image $I^v$ is processed by a frozen MLLM
visual encoder $E_{\mathrm{vis}}(\cdot)$ to obtain a spatial feature map.
Instance-mask-weighted pooling then extracts an object-aligned visual feature:
\begin{equation}
\mathbf{v}_i^{v}
=
\operatorname{MaskPool}
\left(
E_{\mathrm{vis}}(I^{v}),
\mathbf{M}_i^{v}
\right),
\qquad v\in\mathcal{V},
\end{equation}
where $\mathbf{M}_i^{v}$ denotes the instance mask of object $i$ in view $v \in {{\text{top}, \text{oblique}}}$.

We align the object-level SGR feature $\mathbf{z}_i^{\mathrm{SGR}}$ with its
corresponding visual feature $\mathbf{v}_i^v$ using a cosine similarity loss
and an object-level contrastive loss:
\begin{equation}
\mathcal{L}_{\mathrm{vis}}
=
\mathcal{L}_{\mathrm{cos}}
(
\mathbf{z}_i^{\mathrm{long}}, \mathbf{v}_i^v
)
+
\lambda \mathcal{L}_{\mathrm{NCE}}
(
\mathbf{z}_i^{\mathrm{long}}, \mathbf{v}_i^v
),
\end{equation}
where both losses are evaluated over matched object-view pairs.
This alignment is introduced for providing visual supervision during training, while rendered views are not required for online scene understanding.

\subsection{MLLM-guided Semantic Reward Evaluation and Object-level Decision}
\label{sec:reward}

While MLLM is capable of diagnosing the intermediate layout at each denoising step, repeatedly invoking it would incur substantial inference cost. 
To mitigate this issue, we introduce an MLLM-guided Semantic Reward Evaluation and Object-Level Decision module, which distills the MLLM's candidate-level evaluations into a lightweight Reward Predictor.
During inference, the predictor estimates utilities of candidate corrections and selects the correction to guide subsequent denoising.

\noindent\textbf{MLLM-Based Candidate Correction Generation.}
At denoising step $t$, we provide the hierarchical scene representation $\mathbf{U}_t$ and the text prompt $p_s$ to an MLLM, which generates a textual description $\mathcal{T}_t$ of object-level corrections:
\begin{equation}
\mathcal{T}_t = \operatorname{MLLM}(\mathbf{U}_t, p_s).
\end{equation}
We parse $\mathcal{T}_t$ into a structured set of $N_t$ candidate actions,
\begin{equation}
\mathcal{A}_t^{\mathrm{prop}}
= \left\{ a_i^t \right\}_{i=1}^{N_t},
\qquad
a_i^t =
\left(\mathrm{id}_i,\mathbf{M}_i,\mathbf{y}_{i,\mathrm{tar}}\right),
\end{equation}
where $\mathrm{id}_i$ identifies the target object, $\mathbf{M}_i$ specifies the attributes to be modified, and $\mathbf{y}_{i,\mathrm{tar}}$ denotes their
target values. 
In this work, we mainly consider two correction types, horizontal
translation and yaw rotation. 
For the object attributes $(x,y,\psi)$, the corresponding masks are $\mathbf{M}_i=[1,1,0]$ and $\mathbf{M}_i=[0,0,1]$, respectively.
To filter out geometrically invalid proposals, we temporarily apply each candidate action to the current layout and discard it if the resulting configuration violates any scene-level geometric validity constraint.

\noindent\textbf{Semantic Reward Prediction and Object-level Decision.}
We rank the feasible candidate corrections in $\mathcal{A}_t^{\mathrm{prop}}$ using a lightweight Reward Predictor.
The predictor approximates the MLLM-based evaluation of each candidate along three dimensions: Prompt Alignment (P), Functional Coherence (F), and Semantic Consistency (S).

For each feasible action $a$ targeting object $i$, we decode the projected
clean estimate $\tilde{\mathbf{x}}_0^{(t)}$ to obtain the current scene
$\mathcal{S}_t$ and apply $a$ to construct a hypothetical candidate scene
$\mathcal{S}_t^{(a)}$. We then extract the hierarchical representations of
the reference and candidate scenes,
$
\mathbf{U}_t^{\mathrm{ref}}
=
\mathcal{U}(\mathcal{S}_t,p_s)$ and 
$
\mathbf{U}_t^{(a)}
=
\mathcal{U}(\mathcal{S}_t^{(a)},p_s)
$.
% \begin{equation}
% \mathbf{U}_t^{\mathrm{ref}}
% =
% \mathcal{U}(\mathcal{S}_t,p_s),
% \qquad
% \mathbf{U}_t^{(a)}
% =
% \mathcal{U}(\mathcal{S}_t^{(a)},p_s).
% \label{eq:candidate_understanding}
% \end{equation}
Given these representations, the cached prompt feature $\mathbf{q}_{p_s}$, and the action encoding $\iota(a)$, the Reward Predictor $\mathcal{f}_{\eta}(\cdot)$ outputs a score distribution for each evaluation dimension,
\begin{equation}
\left\{
\hat{\boldsymbol{\pi}}_d^{(a)}
\right\}_{d\in\{\mathrm{P,F,S}\}}
=
\mathcal{f}_{\eta}
\left(
\mathbf{U}_t^{\mathrm{ref}},
\mathbf{U}_t^{(a)},
\mathbf{q}_{p_s},
\iota(a)
\right),
\label{eq:reward_prediction}
\end{equation}
where the action encoding $\iota(a)$ specifies the target object, correction type, and proposed target attributes.
Each $\hat{\boldsymbol{\pi}}_d^{(a)}$ is a probability distribution over the five valid ordinal levels $\{0,1,2,3,4\}$ and an additional Unknown class.
The predicted classes and conditional expected scores are then normalized and weighted aggregated with weights of 3:1:1 to calculate a predicted utility score $\hat{u}_a$.

Among these candidates, we select the action with the highest utility after penalizing the magnitude of the proposed correction,
\begin{equation}
\mathcal{D}_t = 
\arg\max_{a\in\mathcal{A}_t^{\mathrm{prop}}}
\hat{u}_a-\lambda_c C_t(a)
\end{equation}
where $C_t(a)$ is the normalized modification cost and $\lambda_c$ controls
its penalty.

We encode each selected decision with the Decision Encoder to obtain a transient conditioning signal for subsequent denoising, 
% $\mathbf{c}_t^{\mathrm{trans}} = E_{\mathrm{dec}}(\mathcal{D}_t)$.
\begin{equation}
\mathbf{c}_t^{\mathrm{trans}}
=
E_{\mathrm{dec}}(\mathcal{D}_t).
\label{eq:transient_condition}
\end{equation}
This signal conveys the selected correction to the diffusion model at the following denoising steps.

At the same time, we collect all accepted corrections within each denoising window of length $\tau_b$ in chronological order:
\begin{equation}
\mathcal{B}_k
=
\operatorname{Buffer}_{\tau_b}
\left(
\{\mathcal{D}_t\}_{t\in\mathcal{W}_k}
\right),
\label{eq:action_buffer}
\end{equation}
where $\mathcal{W}_k$ denotes the $k$-th denoising window. 
% At the end of the window, the buffered corrections are executed sequentially to update the
% persistent correction state. 
% The updated state is then applied through clean-estimate projection to guide subsequent denoising steps.

After each buffering interval $\mathcal{W}_k$ of $\tau_b$ denoising steps, the buffered actions are
sequentially applied to update a persistent correction state, initialized as $\mathcal{P}_0=\varnothing$.
For each corrected object, the state stores its persistent ID, attribute mask, and corresponding
target values for position and yaw. 
The state update is formulated as:
\begin{equation}
\mathcal{P}_{k+1}
=
\operatorname{Update}_{\mathrm{seq}}
(\mathcal{P}_k,\mathcal{B}_k),
\label{eq:persistent_update}
\end{equation}
where $\operatorname{Update}_{\mathrm{seq}}$ sequentially applies buffered actions in chronological order, preserving previously stored targets while allowing later actions to overwrite the same attributes.

\noindent\textbf{MLLM-supervised Reward Distillation.}
During training, a frozen MLLM evaluator $E$ provides
P/F/S score labels, including Unknown, and pairwise
preferences among feasible candidates. The Reward
Predictor $f_\eta$ learns to reproduce individual scores
and relative candidate rankings through:
\begin{equation}
\begin{aligned}
\mathcal{L}_{\mathrm{pref}}
&=\mathcal{L}_{\mathrm{rank}}
+\lambda_{\mathrm{tie}}
\mathbb{E}_{\mathrm{tie}}
[(\hat{u}_a-\hat{u}_b)^2],\\
\mathcal{L}_{R}
&=\mathcal{L}_{\mathrm{score}}^{\mathrm{rows}}
+\mathcal{L}_{\mathrm{score}}^{\mathrm{groups}}\\
&\quad+\lambda_{\mathrm{pref}}\mathcal{L}_{\mathrm{pref}}
+\lambda_{\mathrm{onto}}\mathcal{L}_{\mathrm{ontology}},\\
\mathcal{L}_{\mathrm{score}}
&=\mathcal{L}_{\mathrm{CE}}
+\mathcal{L}_{\mathrm{reg}}
+\lambda_{\mathrm{KL}}\mathcal{L}_{\mathrm{KL}},
\end{aligned}
\label{eq:reward_distillation}
\end{equation}
where score losses supervise individual and group-level evaluations, ranking and tie losses transfer candidate preferences, and the ontology loss regularizes cross-dimensional consistency. 
The $\lambda$ terms control the relative contributions of each objective.
Only $f_\eta$ is optimized, while $E$ and $\mathcal{U}$ remain frozen. 
At inference, $E$ is removed and $f_\eta$ evaluates candidate corrections.

\subsection{Decision-conditioned Trajectory-consistent Diffusion}
\label{sec:diffusion}

To learn localized correction, we construct decision supervision through single-object perturbations.
Given a clean layout $\mathbf{x}_0^*$, we randomly select one object $i$ and perturb its horizontal position or yaw, obtaining $\mathbf{x}_0^c$ while leaving other attributes unchanged. 
The training decision and its encoded condition are:
\begin{equation}
\begin{aligned}
\mathbf{y}_{i,\mathrm{tar}}^*
&=
\mathbf{m}_{\tau}\odot\mathbf{y}_i^*
+
(\mathbf{1}-\mathbf{m}_{\tau})\odot\mathbf{y}_i^c,\\
\mathcal{D}^{\mathrm{train}}
&=
(\mathrm{id}_i,\mathbf{m}_{i},
\mathbf{y}_{i,\mathrm{tar}}^*),\\
\mathbf{c}^{\mathrm{train}}
&=
E_{\mathrm{dec}}(\mathcal{D}^{\mathrm{train}}),
\end{aligned}
\label{eq:training_decision}
\end{equation}
where $\mathbf{y}_i=[p_i^x,p_i^z,\theta_i]$, and $\mathbf{m}_{\tau}$ selects the perturbed attributes.
$\mathbf{y}_i^*$ and $\mathbf{y}_i^c$ denote the clean and corrupted attributes, respectively.

We retain a frozen baseline denoiser
$\mathbf{v}_{\theta_0}$, initialized from the same
pretrained weights as the trainable denoiser
$\mathbf{v}_{\theta}$. The corrupted layout
$\mathbf{x}_0^c$ is forward-diffused with a shared
noise sample to obtain $\mathbf{x}_t^c$.
Both denoisers receive the same noised input, timestep,
prompt, and noise realization. The baseline receives
no decision condition, whereas the trainable denoiser
is conditioned on the encoded training decision.
Their clean estimates are reconstructed as:
\begin{equation}
\begin{aligned}
\hat{\mathbf{x}}_0^{\mathrm{base}}
&=
\sqrt{\bar{\alpha}_t}\mathbf{x}_t^c
-
\sqrt{1-\bar{\alpha}_t}
\mathbf{v}_{\theta_0}
(\mathbf{x}_t^c,t,p_s,\mathbf{c}_{\varnothing}),\\
\hat{\mathbf{x}}_0^{\mathrm{dec}}
&=
\sqrt{\bar{\alpha}_t}\mathbf{x}_t^c
-
\sqrt{1-\bar{\alpha}_t}
\mathbf{v}_{\theta}
(\mathbf{x}_t^c,t,p_s,
E_{\mathrm{dec}}(\mathcal{D}^{\mathrm{train}})).
\end{aligned}
\label{eq:decision_predictions}
\end{equation}

The denoiser and Decision Encoder are optimized with:
\begin{equation}
\begin{aligned}
& \mathcal{L}_{\mathrm{DCD}}
=\mathcal{L}_{\mathrm{diff}}^{\mathrm{clean}}
+\lambda_{\mathrm{target}}\mathcal{L}_{\mathrm{target}}
+\lambda_{\mathrm{preserve}}\mathcal{L}_{\mathrm{preserve}},\\
&\mathcal{L}_{\mathrm{target}}
=
\|\mathbf{M}\odot
(\hat{\mathbf{x}}_0^{\mathrm{dec}}-\mathbf{x}_0^*)\|_1,\\
&\mathcal{L}_{\mathrm{preserve}}
=
\|(1-\mathbf{M})\odot
(\hat{\mathbf{x}}_0^{\mathrm{dec}}
-\operatorname{sg}(\hat{\mathbf{x}}_0^{\mathrm{base}}))\|_1,
\end{aligned}
\label{eq:dcd_objective}
\end{equation}
where $\mathcal{L}_{\mathrm{diff}}^{\mathrm{clean}}$
preserves generation capability, and $\mathbf{M}$
selects the perturbed attributes. The target loss
encourages recovery of the clean attributes, while the
preservation loss distills baseline behavior on
unselected attributes. The two coefficients control
the relative contributions of correction and preservation;
$\operatorname{sg}$ denotes stop-gradient.

At inference, selected decisions are accumulated through buffered sequential execution to update the persistent correction state $\mathcal{P}_k$. 
At each reverse-diffusion step, we enforce the accumulated targets by softly
projecting the predicted clean attributes:
\begin{equation}
\tilde{\mathbf{y}}_i
=
\mathcal{W}\left(
\hat{\mathbf{y}}_i
+
\rho\bar{\mathbf{M}}_{i,k}
\odot
\mathcal{W}\left(
\bar{\mathbf{y}}_{i,\mathrm{tar}}^{(k)}
-\hat{\mathbf{y}}_i
\right)
\right),
\label{eq:clean_projection}
\end{equation}
where $\hat{\mathbf{y}}_i= [\hat{p}_i^x,\hat{p}_i^z,\hat{\theta}_i]$ is extracted from the predicted clean estimate, and $\bar{\mathbf{M}}_{i,k}$ and $\bar{\mathbf{y}}_{i,\mathrm{tar}}^{(k)}$ denote the accumulated attribute mask and target retrieved from $\mathcal{P}_k$, respectively.
$\rho\in(0,1]$ controls the projection strength, and $\mathcal{W}$ wraps only the yaw component while leaving position unchanged. 

The projected attributes are composed back into the original clean estimate, with yaw re-encoded using sine and cosine:
\begin{equation}
\begin{aligned}
\tilde{\mathbf{x}}_0^{(t)}
% &=
% \operatorname{Compose}
% \left(
% \hat{\mathbf{x}}_0^{(t)},
% \left\{
% \tilde{\mathbf{p}}_i^{xz},
% \sin\tilde{\theta}_i,
% \cos\tilde{\theta}_i
% \right\}_{i=1}^{N}
% \right)\\
&=
\mathbf{M}_k^{\mathrm{layout}}
\odot\mathbf{x}_{0,\mathrm{proj}}^{(t)}
+
(1-\mathbf{M}_k^{\mathrm{layout}})
\odot\hat{\mathbf{x}}_0^{(t)},
\end{aligned}
\label{eq:projected_clean}
\end{equation}
where $\mathbf{M}_k^{\mathrm{layout}}$ expands the persistent attribute masks to the corresponding layout channels, including both sine and cosine channels for yaw. 
The resulting $\mathbf{x}_{0,\mathrm{proj}}^{(t)}$ contains the projected target attributes at the selected channels. 
It is then incorporated into the reverse diffusion process:
\begin{equation}
\mathbf{x}_{t-1}
=
\operatorname{DDPM}
\left(
\mathbf{x}_t,
\tilde{\mathbf{x}}_0^{(t)},
t
\right),
\label{eq:ddpm_transition}
\end{equation}
where $\operatorname{DDPM}$ denotes the standard posterior sampling operator, taking the current noisy state, projected clean estimate, and diffusion timestep as inputs to produce the next denoising state. 
By repeatedly enforcing accumulated corrections along the same reverse diffusion trajectory, this mechanism maintains accepted target modifications while preserving non-target scene attributes.
Further training and posterior sampling details are provided in the supplementary material.

\section{Experiment}

\begin{figure*}[!t]
\centering
\includegraphics[width=0.92\linewidth]{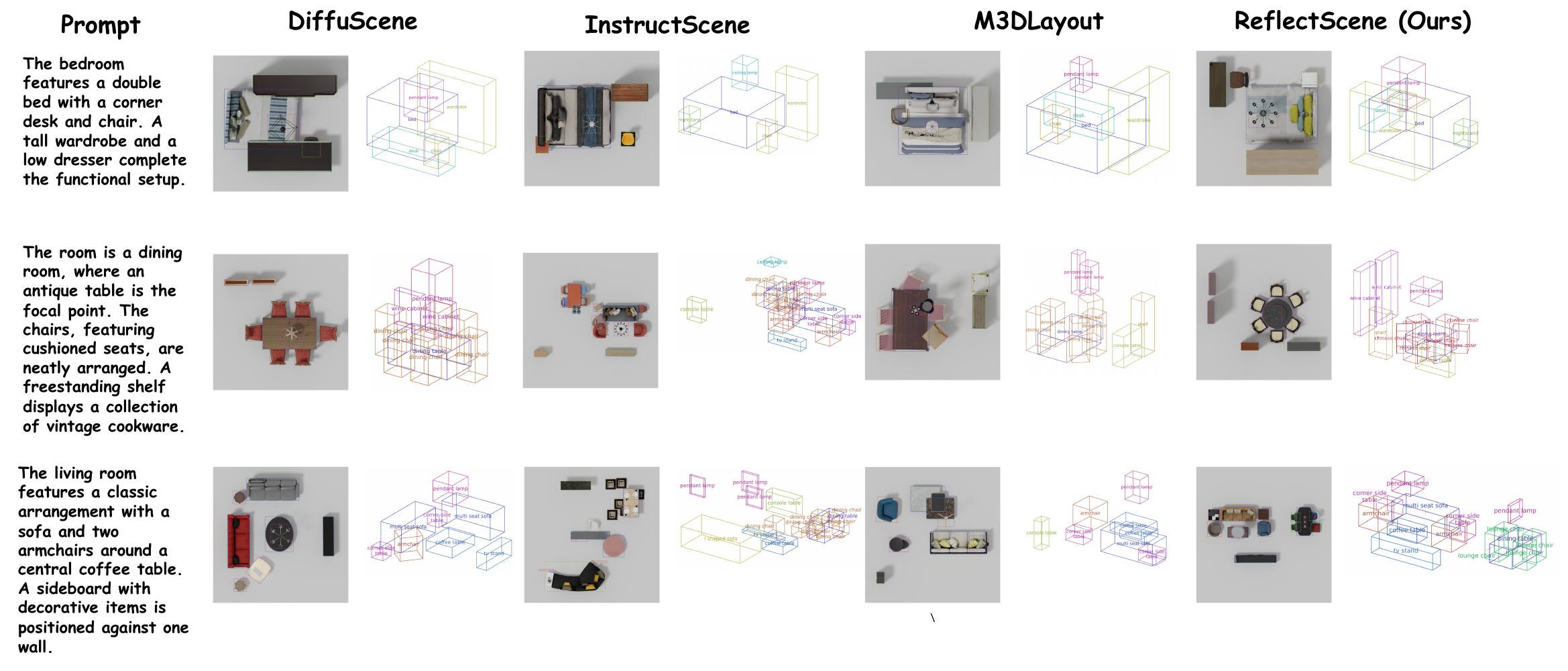}
\vspace{-12pt}
\caption{\textbf{Qualitative comparison.} Across three room types, ReflectScene better preserves prompt-specified furniture arrangements and functional spatial relationships, producing more coherent layouts. Top: rendered scenes; bottom: 3D bounding-box visualizations.}
\label{fig:3}
\vspace{-8pt}
\end{figure*}

\begin{table*}[t]
    \centering
    % \vspace{-10pt}
    \caption{Quantitative comparison of different methods on Bedroom, Dining Room, and Living Room scenes.}
    \label{tab:quantitative_comparison}
    \vspace{-8pt}
    \setlength{\tabcolsep}{3.2pt}
    \renewcommand{\arraystretch}{1.15}
    
    \resizebox{\textwidth}{!}{
    \begin{tabular}{@{}ll*{18}{c}@{}}
        \toprule
        \multirow{2}{*}{\textbf{Method}} &
        \multirow{2}{*}{\textbf{Venue}} &
        \multicolumn{6}{c}{\textbf{Bedroom}} &
        \multicolumn{6}{c}{\textbf{Dining Room}} &
        \multicolumn{6}{c}{\textbf{Living Room}} \\
        
        \cmidrule(lr){3-8}
        \cmidrule(lr){9-14}
        \cmidrule(lr){15-20}
        
        & &
        FID$\downarrow$ & KID$\downarrow$ & SCA\% &  PSA$\uparrow$ & CLIP-Score$\uparrow$ & \shortstack{PIoU$\downarrow$} 
        &
        FID$\downarrow$ & KID$\downarrow$ & SCA\% & PSA$\uparrow$ & CLIP-Score$\uparrow$ & \shortstack{PIoU$\downarrow$}
        &
        FID$\downarrow$ & KID$\downarrow$ & SCA\% & PSA$\uparrow$ & CLIP-Score$\uparrow$ & \shortstack{PIoU$\downarrow$}
        \\
        
        \midrule
        
        DiffuScene\cite{tang2024diffuscene} & CVPR 2024 &
        118.11 & \uline{3.53} & \textbf{62.29} & \uline{74.2} & 0.2457 & \uline{0.77} &
        144.90 & 13.19 & 83.37 & 61.5 &\uline{0.2123} & 0.84 &
        123.94 & \uline{9.54} & 79.29 & 67.6 & 0.2352 & 0.62 \\
        
        InstructScene\cite{lin2024instructscene} & ICLR 2024 &
        131.47 & 8.89 & 71.57 &69.7  & 0.2282 & 0.86 &
        143.02 & 17.98 & 91.22 & 60.8 & 0.2118 & 0.82 &
        127.97 & 10.42 & 83.91 & 62.4 & 0.2237 & 0.47\\
        
        M3DLayout\cite{zhang2026m3dlayout} & CVPR 2026 &
        \uline{115.63} & 5.01 & 64.62 & 73.9 & \uline{0.2463} & 0.78 &
        \uline{138.10} & \uline{12.84} & \textbf{78.46} & \uline{66.4} & 0.2034 & \textbf{0.74} &
        \uline{122.78} & \textbf{9.48} & \uline{77.95} & \uline{71.3} & \uline{0.2366} & \textbf{0.31} \\
        
        \grayrow
        \textbf{ReflectScene} & \textbf{Ours} &
            \textbf{108.70} & \textbf{2.97} & \uline{62.31} & \textbf{82.5} & \textbf{0.2466} & \textbf{0.71}&
        \textbf{117.38} &\textbf{9.28}  & \uline{83.33} & \textbf{74.7}  &\textbf{0.2159}&
        \uline{0.77}&
        \textbf{110.62} &9.63  &\textbf{71.28} & \textbf{78.1}  & \textbf{0.2546} & \uline{0.44} \\
        
        \bottomrule
    \end{tabular}
    }
    \vspace{-12pt}
\end{table*}

\subsection{Datasets and Implementation Details}
\noindent\textbf{Datasets, Rendering, and Evaluation.}
We use 3D-FRONT \cite{Fu_2021_ICCV} (4,110/1,068/576 train/val/test scenes) and 3D-FUTURE \cite{Fu_2021_IJCV} (33 furniture categories). The test set contains 192 scenes per room type (Bedroom, Dining, Living) and 1,500 prompts; comparisons are category-wise, while ablations and capability evaluations use all 576 scenes. Following M3DLayout \cite{zhang2026m3dlayout}, all methods share identical assets and rendering settings. Evaluation covers distributional fidelity (FID/KID), text-scene alignment (CLIP-Score), distribution similarity (SCA; 50\% indicates similarity), spatial overlap (PIoU), and physical-semantic consistency (PSA).

\noindent\textbf{Implementation Details.}
ReflectScene uses M3DLayout on 2$\times$RTX 5090 GPUs. Its 512-channel 1D U-Net (120 slots; frozen BERT) trains for 28K epochs (Adam; batch 256; lr $2\times10^{-4}$; 1K steps). Scene understanding uses a frozen 3B MLLM with two-stage multi-view training. The P evaluation head and Reward Predictor are trained for 2K/2.4K updates (batch 64; lr $10^{-3}$), using MLLM P/F/S scores and preference supervision. DCD fine-tunes the denoiser and Decision Encoder against an identically initialized frozen baseline (2.56K updates; AdamW; batch 64; lr $10^{-5}/10^{-4}$; clean/target/preservation loss weights 1.0/0.4/0.2). Inference performs per-step selection and executes buffered actions every 50 steps via persistent clean-estimate projection.Further details are provided in the supplementary material.

\subsection{Comparisons with Existing Methods}
\noindent\textbf{Quantitative Results.}
We reproduce and retrain competing methods using public implementations under identical data splits and rendering protocols. As shown in \cref{tab:quantitative_comparison}, ReflectScene achieves the best FID, PSA, and CLIP-Score across all room types, reducing FID by 6.0\%, 15.0\%, and 9.9\% over M3DLayout, respectively. It also achieves the lowest KID on Bedroom and Dining Room while maintaining competitive SCA and PIoU, demonstrating improved semantic and functional coherence without compromising spatial plausibility.

\noindent\textbf{Qualitative Results.}As shown in \cref{fig:3}, ReflectScene produces more semantically aligned and functionally coherent layouts across all three room types, better preserving prompt-specified furniture arrangements than competing methods. The rendered scenes and 3D bounding boxes further demonstrate improved text-scene alignment and spatial organization.

\subsection{Ablation Study}

\begin{table}[t]
    \centering
    \caption{Progressive ablation study of the three components in ReflectScene.}
    \label{tab:ablation_components}
    \vspace{-8pt}
    \footnotesize
    \setlength{\tabcolsep}{3pt}
    \renewcommand{\arraystretch}{1.2}

    \resizebox{\columnwidth}{!}{
    \begin{tabular}{@{}lcccccc@{}}
        \toprule
        \textbf{Variant} &
        \textbf{FID}$\downarrow$ &
        \textbf{KID}$\downarrow$ &
        \textbf{SCA\%} &
        \textbf{PSA}$\uparrow$ &
        \textbf{CLIP-Score}$\uparrow$ &
        \textbf{PIoU}$\downarrow$ \\
        \midrule

        \grayrow
        \textbf{Ours}
        & \textbf{69.37} & \textbf{5.23} & \textbf{74.38}& \textbf{78.4} & \textbf{0.2425} & \textbf{0.64} \\
        
        w/o (Reward Evaluation)
        & 71.89 & 6.36 & 76.52 & 75.3 & 0.2412 & 0.65 \\

        \shortstack[l]{w/o (Reward Evaluation\\
        + Scene Understanding)}
        & 75.05 & 6.70 & 77.37 & 71.1 & 0.2395 & 0.71 \\

        \shortstack[l]{w/o (Reward Evaluation\\
        + Scene Understanding + DCD)}
        & 80.22 & 9.75 & 78.91 & 66.7 & 0.2189 & 0.73 \\

        \bottomrule
    \end{tabular}
    }
    \vspace{-12pt}
\end{table}
\noindent\textbf{Main Ablation Study.}
\cref{tab:ablation_components} reports the results of progressively removing the three components to assess their contributions, evaluating all variants on the full test set. 
Removing Reward Evaluation degrades performance, increasing FID from 69.37 to 71.89 and reducing PSA from 78.4 to 75.3. 
Further removing Scene Understanding causes additional declines in generation quality and semantic alignment. Removing DCD leads to the largest degradation, with FID rising to 80.22, PSA dropping to 66.7, and CLIP-Score falling to 0.2189.
Meanwhile, PIoU increases from 0.64 to 0.73, indicating more severe object overlaps, while SCA moves further from the ideal 50\%.
These results confirm the complementary roles of the three components in semantic understanding, corrective decision-making, and preserving corrections throughout denoising.

\noindent\textbf{Reward Predictor.}
As shown in \cref{tab:reward_student_evaluation}, on an independent acceptance set of 40 novel scenes, 93 valid state tasks, and 1,488 questions, the Reward Predictor achieves 98.98\% task-preference accuracy, 97.46\% P-balanced accuracy, and 100\% preference coverage. 
It reduces median candidate-evaluation latency from 506.24 ms to 9.66 ms (52.4$\times$ faster), while supplementary results confirm high P/F/S distillation fidelity with 98.20\% candidate preference agreement. These results validate that MLLM-based semantic evaluation can be effectively distilled into a lightweight predictor, preserving reliable candidate-level decisions while substantially reducing evaluation overhead.

\begin{table}[t]
    \centering
    \caption{Teacher--student comparison for candidate-level reward evaluation on an independent acceptance set of 40 novel scenes, 93 valid state tasks, and 1,488 questions.}
    \label{tab:reward_student_evaluation}
    \vspace{-8pt}
    \footnotesize
    \setlength{\tabcolsep}{3pt}
    \renewcommand{\arraystretch}{1.2}
    \begin{tabularx}{\columnwidth}{
        @{}
        >{\raggedright\arraybackslash}X
        >{\centering\arraybackslash}p{0.27\columnwidth}
        >{\centering\arraybackslash}p{0.27\columnwidth}
        @{}
    }
        \toprule
        \textbf{Metric} &
        \textbf{\shortstack{Frozen MLLM\\Evaluator (Teacher)}} &
        \textbf{\shortstack{Reward Predictor\\(Student)}} \\
        \midrule

        Task-Preference Accuracy ($\uparrow$)
        & \textbf{99.24\%} & 98.98\% \\

        Preference Coverage ($\uparrow$)
        & 99.90\% & \textbf{100\%} \\

        P-Balanced Accuracy ($\uparrow$)
        & \textbf{98.26\%} & 97.46\% \\

        Missing-Evidence Unknown Recall ($\uparrow$)
        & \textbf{100\%} & 99.73\% \\

        Known-Score Preservation Rate ($\uparrow$)
        & \textbf{100\%} & \textbf{100\%} \\

        Median Candidate-Evaluation Latency ($\downarrow$)
        & 506.24\,ms & \textbf{9.66\,ms} \\

        \bottomrule
    \end{tabularx}
    \vspace{2pt}
    \begin{minipage}{\columnwidth}
        \scriptsize
        Latency is measured for evaluating five fixed candidates per state item, with request features cached. Both methods exclude upstream scene understanding (U) computation. Preference accuracy should be interpreted together with preference coverage.
    \end{minipage}
    \vspace{-20pt}
\end{table}

\noindent\textbf{SGR Adapter and Multi-view Alignment.}
As shown in \cref{tab:adapter_mllm_comparison}, integrating the frozen MLLM improves semantic, relational, and functional understanding over Adapter-only, demonstrating the value of contextual reasoning. As shown in \cref{tab:state_understanding_alignment}, the SGR Adapter maintains above 91\% Macro-F1 and below 1.4 cm XZ error across DDPM stages, while visual matching recovers from 59.33\% to 73.16\%, approaching the clean-layout result of 73.26\%. 
Additional supplementary evaluations confirm its fine-grained understanding, cross-seed stability, and counterfactual robustness, validating reliable scene understanding and visual grounding throughout denoising.

\begin{table}[t]
    \centering
    % \vspace{-10pt}
    \caption{Comparison of adapter-only tuning and adapter tuning with a frozen MLLM for scene understanding.}
    \vspace{-8pt}
    \label{tab:adapter_mllm_comparison}
    \scriptsize
    \setlength{\tabcolsep}{2pt}
    \renewcommand{\arraystretch}{1.05}
    \begin{tabularx}{\columnwidth}{
        @{}
        >{\raggedright\arraybackslash}X
        >{\centering\arraybackslash}p{0.23\columnwidth}
        >{\centering\arraybackslash}p{0.27\columnwidth}
        @{}
    }
        \toprule
        \textbf{Metric} &
        \textbf{Adapter-only} &
        \textbf{\shortstack{Adapter +\\Frozen MLLM}} \\
        \midrule
        Category Macro-F1 ($\uparrow$)
        & 84.58\% & \textbf{96.68\%} \\

        Distance Rel. F1 ($\uparrow$)
        & 82.25\% & \textbf{94.07\%} \\

        Adjacency Rel. F1 ($\uparrow$)
        & 71.92\% & \textbf{82.82\%} \\

        Weak Functional Role F1 ($\uparrow$)
        & 90.17\% & \textbf{99.11\%} \\

        Room Category Accuracy ($\uparrow$)
        & 86.77\% & \textbf{90.77\%} \\

        Activity Binary F1 ($\uparrow$)
        & 93.81\% & \textbf{98.73\%} \\

        XZ Center MAE ($\downarrow$)
        & 1.57\,cm & \textbf{1.21\,cm} \\

        Yaw MAE ($\downarrow$)
        & 0.98$^\circ$ & \textbf{0.91$^\circ$} \\
        \bottomrule
    \end{tabularx}
    \vspace{-8pt}
\end{table}

\begin{table}[t]
    \centering
    \caption{Scene understanding and object-visual alignment across diffusion states. Cat. F1 and Dist. F1 denote category and distance-relation Macro-F1, respectively; XZ MAE denotes the XZ center mean absolute error, and Visual Top-1 denotes object-visual matching Top-1 accuracy.}
    \label{tab:state_understanding_alignment}
    \vspace{-8pt}
    \scriptsize
    \setlength{\tabcolsep}{2pt}
    \renewcommand{\arraystretch}{1.05}
    \begin{tabularx}{\columnwidth}{
        @{}
        >{\raggedright\arraybackslash}p{0.20\columnwidth}
        >{\centering\arraybackslash}p{0.15\columnwidth}
        >{\centering\arraybackslash}p{0.15\columnwidth}
        >{\centering\arraybackslash}p{0.14\columnwidth}
        >{\centering\arraybackslash}X
        @{}
    }
        \toprule
        \textbf{State} &
        \textbf{\shortstack{Cat. F1\\($\uparrow$)}} &
        \textbf{\shortstack{Dist. F1\\($\uparrow$)}} &
        \textbf{\shortstack{XZ MAE\\($\downarrow$)}} &
        \textbf{\shortstack{Visual Top-1\\($\uparrow$)}} \\
        \midrule

        Clean
        & 98.28\% & 95.48\% & 1.15\,cm & 73.26\% \\

        Perturbed
        & 98.04\% & 95.36\% & 1.23\,cm & 70.34\% \\

        DDPM Early
        & 94.43\% & 93.29\% & 1.36\,cm & 59.33\% \\

        DDPM Middle
        & 95.65\% & 92.30\% & 1.22\,cm & 70.51\% \\

        DDPM Late
        & 95.21\% & 91.43\% & 1.14\,cm & 73.16\% \\

        \midrule
        Overall
        & -- & -- & -- & \textbf{70.38\%} \\

        \bottomrule
    \end{tabularx}
    \vspace{-8pt}
\end{table}

\subsection{Models with Different Capabilities}
\cref{tab:different_model} compares different reasoning strategies and MLLM backbones to investigate their impact on scene generation. 
Adapter-only and Geometric Algorithm achieve comparable performance to Qwen2-VL-2B, suggesting that bypassing contextual MLLM reasoning or relying on geometric heuristics provides limited benefits.
Increasing backbone capability to Qwen2.5-VL-3B and 7B yields substantial improvements across all metrics, highlighting the importance of effective semantic and functional reasoning.
However, scaling further to Qwen3-VL-32B brings only marginal gains over 7B, indicating diminishing returns as stronger reasoning capabilities approach the demands of structured indoor scene generation. 
This suggests that effective semantic reasoning, rather than model scale alone, is crucial for improving scene generation.

\begin{table}[t]
    \centering
    \caption{Quantitative comparison of different scene reasoning strategies and MLLM backbones.}
    \vspace{-8pt}
    \label{tab:different_model}
    \footnotesize
    \setlength{\tabcolsep}{3pt}
    \renewcommand{\arraystretch}{1.2}

    \resizebox{\columnwidth}{!}{
    \begin{tabular}{@{}lcccccc@{}}
        \toprule
        \textbf{Model} &
        \textbf{FID}$\downarrow$ &
        \textbf{KID}$\downarrow$ &
        \textbf{SCA\%} &
        \textbf{PSA}$\uparrow$ &
        \textbf{CLIP-Score}$\uparrow$ &
        \textbf{PIoU}$\downarrow$ \\
        \midrule

        Adapter-only
        &76.04  &10.95  &83.36  &65.8  &0.2024  &0.77  \\
        
        Geometric Algorithm
        &78.33  &9.52  &81.29  &69.1  &0.2116  &0.74  \\
        
        Qwen2-VL-2B
        &78.66  &9.43  &82.71  &70.2  &0.2133  &0.74  \\

        Qwen2.5-VL-3B
        &69.37 & 5.23 & 74.38& 78.4 & 0.2425 & 0.64 \\

        Qwen2.5-VL-7B
        &63.71  &3.57  &70.12  &83.1  &0.2610  &0.58  \\

        Qwen3‑VL‑32B
        &62.98  &3.24  &69.86  &84.5  &0.2671  &0.56  \\

        \bottomrule
    \end{tabular}
    }
    \vspace{-20pt}
\end{table}

\section{Conclusion}
We presented ReflectScene, an MLLM-guided closed-loop
diffusion framework integrating hierarchical scene
understanding, reward-driven decisions, and
decision-conditioned denoising to guide scene evolution
while preserving corrections. Experiments on 3D-FRONT
show improved functional consistency and semantic
alignment with competitive generation quality and diversity.

\section*{LLM Usage Statement}

During the preparation of this work the author(s) used ChatGPT in order to check grammar mistakes. 
After using this tool/service, the author(s) reviewed and edited the content as needed and take(s) full responsibility for the content of the published article.

{
    \small
    \bibliographystyle{ieeenat_fullname}
    \bibliography{main}

@String(CVPR= {IEEE Conf. Comput. Vis. Pattern Recog.})

@String(ICCV= {Int. Conf. Comput. Vis.})

@String(TOG= {ACM Trans. Graph.})

@String(AAAI = {AAAI})

@String(CVPR  = {CVPR})

@String(ICCV  = {ICCV})

@String(TOG   = {ACM TOG})

@inproceedings{tang2024diffuscene,
  title={Diffuscene: Denoising diffusion models for generative indoor scene synthesis},
  author={Tang, Jiapeng and Nie, Yinyu and Markhasin, Lev and Dai, Angela and Thies, Justus and Nie{\ss}ner, Matthias},
  booktitle={2024 IEEE/CVF Conference on Computer Vision and Pattern Recognition (CVPR)},
  pages={20507--20518},
  year={2024},
  organization={IEEE}
}

@article{paschalidou2021atiss,
  title={Atiss: Autoregressive transformers for indoor scene synthesis},
  author={Paschalidou, Despoina and Kar, Amlan and Shugrina, Maria and Kreis, Karsten and Geiger, Andreas and Fidler, Sanja},
  journal={Advances in neural information processing systems},
  volume={34},
  pages={12013--12026},
  year={2021}
}

@inproceedings{lin2024instructscene,
  title={Instructscene: Instruction-driven 3d indoor scene synthesis with semantic graph prior},
  author={Lin, Chenguo and Mu, Yadong},
  booktitle={International Conference on Learning Representations},
  volume={2024},
  pages={25687--25718},
  year={2024}
}

@inproceedings{yang2025mmgdreamer,
  title={Mmgdreamer: Mixed-modality graph for geometry-controllable 3d indoor scene generation},
  author={Yang, Zhifei and Lu, Keyang and Zhang, Chao and Qi, Jiaxing and Jiang, Hanqi and Ma, Ruifei and Yin, Shenglin and Xu, Yifan and Xing, Mingzhe and Xiao, Zhen and others},
  booktitle={Proceedings of the AAAI conference on artificial intelligence},
  volume={39},
  number={9},
  pages={9391--9399},
  year={2025}
}

@inproceedings{sun2026semlayoutdiff,
  title={Semlayoutdiff: Semantic layout generation with diffusion model for indoor scene synthesis},
  author={Sun, Xiaohao and Goel, Divyam and Chang, Angel X},
  booktitle={2026 International Conference on 3D Vision (3DV)},
  pages={1545--1555},
  year={2026},
  organization={IEEE}
}

@inproceedings{xu2024pointllm,
  title={Pointllm: Empowering large language models to understand point clouds},
  author={Xu, Runsen and Wang, Xiaolong and Wang, Tai and Chen, Yilun and Pang, Jiangmiao and Lin, Dahua},
  booktitle={European Conference on Computer Vision},
  pages={131--147},
  year={2024},
  organization={Springer}
}

@inproceedings{deng20253d,
  title={3d-llava: Towards generalist 3d lmms with omni superpoint transformer},
  author={Deng, Jiajun and He, Tianyu and Jiang, Li and Wang, Tianyu and Dayoub, Feras and Reid, Ian},
  booktitle={2025 IEEE/CVF Conference on Computer Vision and Pattern Recognition (CVPR)},
  pages={3772--3782},
  year={2025},
  organization={IEEE}
}

@article{pan2026repfusion,
  title={RepFusion: Leveraging Multimodal Priors for Denoising in Representation Space},
  author={Pan, Xichen and Singh, Aashu and Shukla, Satya Narayan and Fan, Xiangjun and Mishra, Shlok Kumar and Xie, Saining},
  journal={arXiv preprint arXiv:2606.14700},
  year={2026}
}

@inproceedings{yang2024physcene,
  title={Physcene: Physically interactable 3d scene synthesis for embodied ai},
  author={Yang, Yandan and Jia, Baoxiong and Zhi, Peiyuan and Huang, Siyuan},
  booktitle={2024 IEEE/CVF Conference on Computer Vision and Pattern Recognition (CVPR)},
  pages={16262--16272},
  year={2024},
  organization={IEEE}
}

@article{feng2026sdesc3d,
  title={SDesc3D: Towards Layout-Aware 3D Indoor Scene Generation from Short Descriptions},
  author={Feng, Jie and Shen, Jiawei and Huang, Junjia and Zhang, Junpeng and Feng, Mingtao and Dong, Weisheng and Li, Guanbin},
  journal={arXiv preprint arXiv:2604.01972},
  year={2026}
}

@article{zhao2026scenerevis,
  title={SceneReVis: A Self-Reflective Vision-Grounded Framework for 3D Indoor Scene Synthesis via Multi-turn RL},
  author={Zhao, Yang and Sun, Shizhao and Zhang, Meisheng and Shi, Yingdong and Yang, Xubo and Bian, Jiang},
  journal={arXiv preprint arXiv:2602.09432},
  year={2026}
}

@article{wang2026function2scene,
  title={Function2scene: 3d indoor scene layout from functional specifications},
  author={Wang, Ruiqi and Chen, Qimin and Ritchie, Daniel and Chang, Angel X and Savva, Manolis and Wang, Kai and Zhang, Hao},
  journal={arXiv preprint arXiv:2605.30819},
  volume={3},
  year={2026}
}

@article{li2019grains,
  title={Grains: Generative recursive autoencoders for indoor scenes},
  author={Li, Manyi and Patil, Akshay Gadi and Xu, Kai and Chaudhuri, Siddhartha and Khan, Owais and Shamir, Ariel and Tu, Changhe and Chen, Baoquan and Cohen-Or, Daniel and Zhang, Hao},
  journal={ACM Transactions on Graphics (TOG)},
  volume={38},
  number={2},
  pages={1--16},
  year={2019},
  publisher={ACM, New York, NY, USA}
}

@inproceedings{zhou2019scenegraphnet,
  title={Scenegraphnet: Neural message passing for 3d indoor scene augmentation},
  author={Zhou, Yang and While, Zachary and Kalogerakis, Evangelos},
  booktitle={2019 IEEE/CVF International Conference on Computer Vision (ICCV)},
  pages={7383--7391},
  year={2019},
  organization={IEEE}
}

@inproceedings{wang2021sceneformer,
  title={Sceneformer: Indoor scene generation with transformers},
  author={Wang, Xinpeng and Yeshwanth, Chandan and Nie{\ss}ner, Matthias},
  booktitle={2021 International conference on 3D vision (3DV)},
  pages={106--115},
  year={2021},
  organization={IEEE}
}

@article{feng2023layoutgpt,
  title={Layoutgpt: Compositional visual planning and generation with large language models},
  author={Feng, Weixi and Zhu, Wanrong and Fu, Tsu-jui and Jampani, Varun and Akula, Arjun and He, Xuehai and Basu, Sugato and Wang, Xin Eric and Wang, William Yang},
  journal={Advances in neural information processing systems},
  volume={36},
  pages={18225--18250},
  year={2023}
}

@inproceedings{zhu20233d,
  title={3d-vista: Pre-trained transformer for 3d vision and text alignment},
  author={Zhu, Ziyu and Ma, Xiaojian and Chen, Yixin and Deng, Zhidong and Huang, Siyuan and Li, Qing},
  booktitle={2023 IEEE/CVF International Conference on Computer Vision (ICCV)},
  pages={2899--2909},
  year={2023},
  organization={IEEE}
}

@inproceedings{jia2024sceneverse,
  title={Sceneverse: Scaling 3d vision-language learning for grounded scene understanding},
  author={Jia, Baoxiong and Chen, Yixin and Yu, Huangyue and Wang, Yan and Niu, Xuesong and Liu, Tengyu and Li, Qing and Huang, Siyuan},
  booktitle={European Conference on Computer Vision},
  pages={289--310},
  year={2024},
  organization={Springer}
}

@inproceedings{chen2024ll3da,
  title={Ll3da: Visual interactive instruction tuning for omni-3d understanding, reasoning, and planning},
  author={Chen, Sijin and Chen, Xin and Zhang, Chi and Li, Mingsheng and Yu, Gang and Fei, Hao and Zhu, Hongyuan and Fan, Jiayuan and Chen, Tao},
  booktitle={2024 IEEE/CVF Conference on Computer Vision and Pattern Recognition (CVPR)},
  pages={26418--26428},
  year={2024},
  organization={IEEE}
}

@InProceedings{Fu_2021_ICCV,
    author    = {Fu, Huan and Cai, Bowen and Gao, Lin and Zhang, Ling-Xiao and Wang, Jiaming and Li, Cao and Zeng, Qixun and Sun, Chengyue and Jia, Rongfei and Zhao, Binqiang and Zhang, Hao},
    title     = {3D-FRONT: 3D Furnished Rooms With layOuts and semaNTics},
    booktitle = {Proceedings of the IEEE/CVF International Conference on Computer Vision (ICCV)},
    month     = {October},
    year      = {2021},
    pages     = {10933--10942}
}

@article{Fu_2021_IJCV,
    author  = {Fu, Huan and Jia, Rongfei and Gao, Lin and Gong, Mingming and Zhao, Binqiang and Maybank, Stephen J. and Tao, Dacheng},
    title   = {{3D-FUTURE: 3D Furniture Shape with TextURE}},
    journal = {International Journal of Computer Vision},
    volume  = {129},
    number  = {12},
    pages   = {3313--3337},
    year    = {2021},
    doi     = {10.1007/s11263-021-01534-z}
}

@inproceedings{zhang2026m3dlayout,
    author    = {Zhang, Yiheng and
                 Cai, Zhuojiang and
                 Wang, Mingdao and
                 Guo, Meitong and
                 Li, Tianxiao and
                 Lin, Li and
                 Wang, Yuwang},
    title     = {{M3DLayout}: A Multi-Source Dataset of {3D} Indoor Layouts and Structured Descriptions for {3D} Generation},
    booktitle = {Proceedings of the IEEE/CVF Conference on Computer Vision and Pattern Recognition (CVPR)},
    month     = {June},
    year      = {2026},
    pages     = {34217--34226}
}

@article{zhai2023commonscenes,
  title={Commonscenes: Generating commonsense 3d indoor scenes with scene graph diffusion},
  author={Zhai, Guangyao and {\"O}rnek, Evin P{\i}nar and Wu, Shun-Cheng and Di, Yan and Tombari, Federico and Navab, Nassir and Busam, Benjamin},
  journal={Advances in Neural Information Processing Systems},
  volume={36},
  pages={30026--30038},
  year={2023}
}

@inproceedings{huang2023diffusion,
  title={Diffusion-based generation, optimization, and planning in 3d scenes},
  author={Huang, Siyuan and Wang, Zan and Li, Puhao and Jia, Baoxiong and Liu, Tengyu and Zhu, Yixin and Liang, Wei and Zhu, Song-Chun},
  booktitle={2023 IEEE/CVF Conference on Computer Vision and Pattern Recognition (CVPR)},
  pages={16750--16761},
  year={2023},
  organization={IEEE}
}

@inproceedings{feng2025casagpt,
  title={CasaGPT: cuboid arrangement and scene assembly for interior design},
  author={Feng, Weitao and Zhou, Hang and Liao, Jing and Cheng, Li and Zhou, Wenbo},
  booktitle={2025 IEEE/CVF Conference on Computer Vision and Pattern Recognition (CVPR)},
  pages={29173--29182},
  year={2025},
  organization={IEEE}
}

@inproceedings{yang2024holodeck,
  title={Holodeck: Language guided generation of 3d embodied ai environments},
  author={Yang, Yue and Sun, Fan-Yun and Weihs, Luca and VanderBilt, Eli and Herrasti, Alvaro and Han, Winson and Wu, Jiajun and Haber, Nick and Krishna, Ranjay and Liu, Lingjie and others},
  booktitle={2024 IEEE/CVF Conference on Computer Vision and Pattern Recognition (CVPR)},
  pages={16277--16287},
  year={2024},
  organization={IEEE}
}

@inproceedings{miao2026nestwork,
  title={Nestwork: Conditional 3D Furnished House Layout Generation through Latent Heterogeneous Graph Diffusion},
  author={Miao, Shuhan and Cao, Biru and Zhuang, Junling},
  booktitle={Proceedings of the IEEE/CVF Conference on Computer Vision and Pattern Recognition},
  pages={27093--27103},
  year={2026}
}

@inproceedings{ran2025directlayout,
    author    = {Ran, Xingjian and Li, Yixuan and Xu, Linning and Yu, Mulin and Dai, Bo},
    title     = {Direct Numerical Layout Generation for 3D Indoor Scene Synthesis via Spatial Reasoning},
    booktitle = {Advances in Neural Information Processing Systems},
    volume    = {38},
    year      = {2025}
}

@inproceedings{deng2025global,
  title={Global-local tree search in vlms for 3d indoor scene generation},
  author={Deng, Wei and Qi, Mengshi and Ma, Huadong},
  booktitle={2025 IEEE/CVF Conference on Computer Vision and Pattern Recognition (CVPR)},
  pages={8975--8984},
  year={2025},
  organization={IEEE}
}

@inproceedings{xue2023ulip,
  title={Ulip: Learning a unified representation of language, images, and point clouds for 3d understanding},
  author={Xue, Le and Gao, Mingfei and Xing, Chen and Mart{\'\i}n-Mart{\'\i}n, Roberto and Wu, Jiajun and Xiong, Caiming and Xu, Ran and Niebles, Juan Carlos and Savarese, Silvio},
  booktitle={2023 IEEE/CVF Conference on Computer Vision and Pattern Recognition (CVPR)},
  pages={1179--1189},
  year={2023},
  organization={IEEE}
}

@inproceedings{peng2023openscene,
  title={Openscene: 3d scene understanding with open vocabularies},
  author={Peng, Songyou and Genova, Kyle and Jiang, Chiyu and Tagliasacchi, Andrea and Pollefeys, Marc and Funkhouser, Thomas},
  booktitle={2023 IEEE/CVF Conference on Computer Vision and Pattern Recognition (CVPR)},
  pages={815--824},
  year={2023},
  organization={IEEE}
}

@article{jatavallabhula2023conceptfusion,
  title={Conceptfusion: Open-set multimodal 3d mapping},
  author={Jatavallabhula, Krishna Murthy and Kuwajerwala, Alihusein and Gu, Qiao and Omama, Mohd and Chen, Tao and Maalouf, Alaa and Li, Shuang and Iyer, Ganesh and Saryazdi, Soroush and Keetha, Nikhil and others},
  journal={arXiv preprint arXiv:2302.07241},
  year={2023}
}

@inproceedings{feng2026lavigen,
    author    = {Feng, Haoran and Niu, Yifan and Huang, Zehuan and Sun, Yang-Tian and Guo, Chunchao and Peng, Yuxin and Sheng, Lu},
    title     = {Repurposing 3D Generative Model for Autoregressive Layout Generation},
    booktitle = {Proceedings of the IEEE/CVF Conference on Computer Vision and Pattern Recognition (CVPR)},
    month     = {June},
    year      = {2026},
    pages     = {3231--3243}
}

@article{takmaz2023openmask3d,
  title={Openmask3d: Open-vocabulary 3d instance segmentation},
  author={Takmaz, Ay{\c{c}}a and Fedele, Elisabetta and Sumner, Robert W and Pollefeys, Marc and Tombari, Federico and Engelmann, Francis},
  journal={arXiv preprint arXiv:2306.13631},
  year={2023}
}

@inproceedings{chen2020scanrefer,
  title={Scanrefer: 3d object localization in rgb-d scans using natural language},
  author={Chen, Dave Zhenyu and Chang, Angel X and Nie{\ss}ner, Matthias},
  booktitle={European conference on computer vision},
  pages={202--221},
  year={2020},
  organization={Springer}
}

@inproceedings{chen2021scan2cap,
  title={Scan2cap: Context-aware dense captioning in rgb-d scans},
  author={Chen, Zhenyu and Gholami, Ali and Nie{\ss}ner, Matthias and Chang, Angel X},
  booktitle={Proceedings of the IEEE/CVF conference on computer vision and pattern recognition},
  pages={3193--3203},
  year={2021}
}

@inproceedings{azuma2022scanqa,
  title={Scanqa: 3d question answering for spatial scene understanding},
  author={Azuma, Daichi and Miyanishi, Taiki and Kurita, Shuhei and Kawanabe, Motoaki},
  booktitle={2022 IEEE/CVF Conference on Computer Vision and Pattern Recognition (CVPR)},
  pages={19107--19117},
  year={2022},
  organization={IEEE}
}

@inproceedings{sun2025layoutvlm,
  title={Layoutvlm: Differentiable optimization of 3d layout via vision-language models},
  author={Sun, Fan-Yun and Liu, Weiyu and Gu, Siyi and Lim, Dylan and Bhat, Goutam and Tombari, Federico and Li, Manling and Haber, Nick and Wu, Jiajun},
  booktitle={2025 IEEE/CVF Conference on Computer Vision and Pattern Recognition (CVPR)},
  pages={29469--29478},
  year={2025},
  organization={IEEE}
}

@inproceedings{jiang2026hoglayout, 
   author = {Jiang, Haiyan and Zhang, Deyu and Weng, Dongdong and Song, Weitao and Duh, Henry Been-Lirn}, title = {HOG-Layout: Hierarchical 3D Scene Generation, Optimization and Editing via Vision-Language Models}, booktitle = {Proceedings of the IEEE/CVF Conference on Computer Vision and Pattern Recognition (CVPR)}, 
   month = {June}, 
   year = {2026}, 
   pages = {31586-31596} 
}

@inproceedings{kuang2026vulcan,
  title={VULCAN: Tool-Augmented Multi Agents for Iterative 3D Object Arrangement},
  author={Kuang, Zhengfei and Lin, Rui and Zhao, Long and Wetzstein, Gordon and Xie, Saining and Woo, Sanghyun},
  booktitle={Proceedings of the IEEE/CVF Conference on Computer Vision and Pattern Recognition},
  pages={23763--23773},
  year={2026}
}
}


\end{document}